\documentclass[10pt,conference,a4paper]{IEEEtran}
\usepackage{url}
\usepackage{ifpdf}

\ifCLASSINFOpdf
  \usepackage[pdftex]{graphicx}
  \graphicspath{{../pdf/}{../jpeg/}}
  \DeclareGraphicsExtensions{.pdf,.jpeg,.png}
\else
  \usepackage[dvips]{graphicx}
  \graphicspath{{../eps/}}
  \DeclareGraphicsExtensions{.eps}
  \fi
\usepackage{amsmath,amsxtra,amssymb,amsthm,latexsym,amscd,amsfonts}
\usepackage[utf8]{vntex}
\usepackage{tabularx}
\ifCLASSINFOpdf
\usepackage[pdftex]{graphicx}
\DeclareGraphicsExtensions{.pdf,.jpeg,.png,.jpg}
\else
\usepackage[dvips]{graphicx}
\DeclareGraphicsExtensions{.eps}
\fi
\usepackage{array}
\usepackage{epstopdf}
\usepackage{anysize}
\marginsize{2.4cm}{2.0cm}{3.2cm}{3.0cm}
\usepackage{multicol}
\usepackage{balance}
\usepackage{graphicx}
\usepackage{caption}
\usepackage{subcaption}

\makeatletter
\def\ScaleIfNeeded{\ifdim\Gin@nat@width>\linewidth\linewidth\else\Gin@nat@width\fi}

\begin{document}
\columnsep=0.63cm
\def\mathbi#1{\boldsymbol{#1}}
\def\erfc{\:\mathrm{erfc}}
\def\arg{\:\mathrm{arg}}
\def\E{\:\mathrm{E}}
\def\sinc{\:\mathrm{sinc}}
\def\T{\mathrm{T}}
\def\H{\mathrm{H}}
\newcommand{\bigsize}{\fontsize{16pt}{20pt}\selectfont}

%
\include{Abbr}
\title{Xây dựng hệ thống  định vị và điều hướng trong nhà dựa trên monocular SLAM cho Robot di động}

\author{
\IEEEauthorblockN{
Nguyễn Cảnh Thanh, Đỗ  Đức Mạnh và Hoàng Văn Xiêm
} 
\IEEEauthorblockA{ Bộ môn Kỹ thuật Robot, Khoa Điện tử - Viễn Thông \\ Trường Đại học Công Nghệ - Đại học Quốc gia Hà Nội\\
		Email: canhthanhlt@gmail.com, ddmanh99@gmail.com, xiemhoang@vnu.edu.vn}
}
\maketitle

\begin{abstract}
Định vị và điều hướng cho robot là hai vấn đề quan trọng trong robot di động. Trong bài báo đề xuất một hướng tiếp cận cho hệ thống định vị và điều hướng đối với robot hai bánh vi sai dựa trên monocular SLAM. Hệ thống được thực hiện trên hệ điều hành lập trình cho robot (Robot Operating System - ROS). Phần cứng của robot là một robot hai bánh với nền tảng máy tính nhúng Jetson Xavier AGX, camera 2D và một cảm biến LiDAR nhằm thu thập thông tin từ môi trường bên ngoài. Thuật toán A$^*$ và phương pháp tiếp cận cửa sổ (DWA) được áp dụng trong lập kế hoạch đường đi dựa trên bản đồ lưới 2D. Thuật toán ORB\_SLAM3 trích xuất các đặc trưng của môi trường từ đó cung cấp tư thế robot cho qua trình định vị và điều hướng. Cuối cùng, hệ thống được thử nghiệm trong môi trường mô phỏng Gazebo và trực quan hóa qua Rviz từ đó chứng minh độ hiệu quả và tiềm năng của hệ thống trong việc định vị và điều hướng cho robot di động trong nhà.
\end{abstract}

\IEEEpeerreviewmaketitle
\section{GIỚI THIỆU}
\label{Sec:intro}
\subsection{Bối cảnh và động lực}
Ngày nay, Robot di động đã thu hút được nhiều sự chú ý cùng với đó, vấn đề định vị và điều hướng là các công nghệ lõi, dần trở thành một phần không thể thiếu trong robot. Vấn đề định vị và điều hướng được tóm tắt trong ba câu hỏi: "Tôi đang ở đâu?", "Tôi đang đi đâu" và "Làm sao để tôi đi đến đó?" \cite{Khatib2015}. Độ chính xác của định vị ảnh hưởng trực tiếp tới quá trình điều hướng, vì vậy một robot di động có thể thực hiện nhiệm vụ một cách nhanh chóng và chính xác dựa trên khả năng thực hiện của hệ thống định vị \cite{Zhang2015}.

Trong hệ thống điều hướng của robot, các cảm biến dựa trên dead-reckoning (DR) được sử dụng rộng rãi \cite{Khatib2015} như Odometry, IMU. Phương pháp DR ước tính tổng quãng đường di chuyển từ điểm xuất phát tuy nhiên sai số ước lượng sẽ được tích lũy theo thời gian. Để cải thiện độ chính xác của quá trình định vị, các loại cảm biến hiện đại được tích hợp như sonar, beacon, camera, laser, GPS, IPS, .. Bên cạnh đó, các điểm đánh dấu cũng ngày càng sử dụng rộng rãi bao gồm các mốc tự nhiên và mốc nhân tạo. Các bộ lọc Bayesian dựa trên các thuật toán xác suất như Kalman và Particle \cite{Liu2016} được sử dụng nhằm ước tính trạng thái của hệ thống từ thông tin của cảm biến như chuỗi Markov. 

\begin{figure}[!ht]
    \centering
    \includegraphics[width=0.48\textwidth]{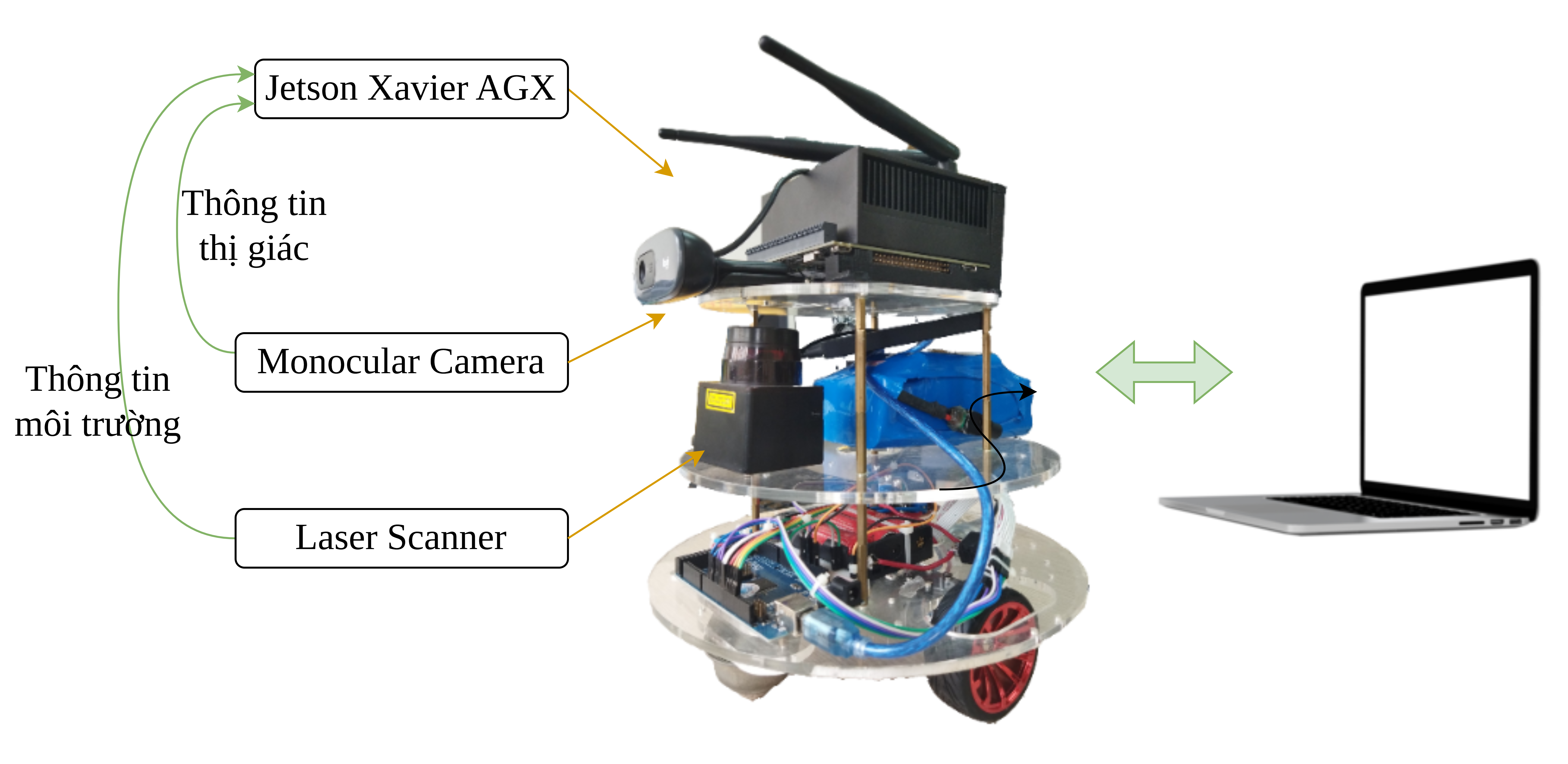}
    \caption{Tổng quan hệ thống phần cứng}
    \label{fig:system_overview}
\end{figure}

\begin{figure*}[!ht]
    \centering
    \includegraphics[width=0.85\textwidth]{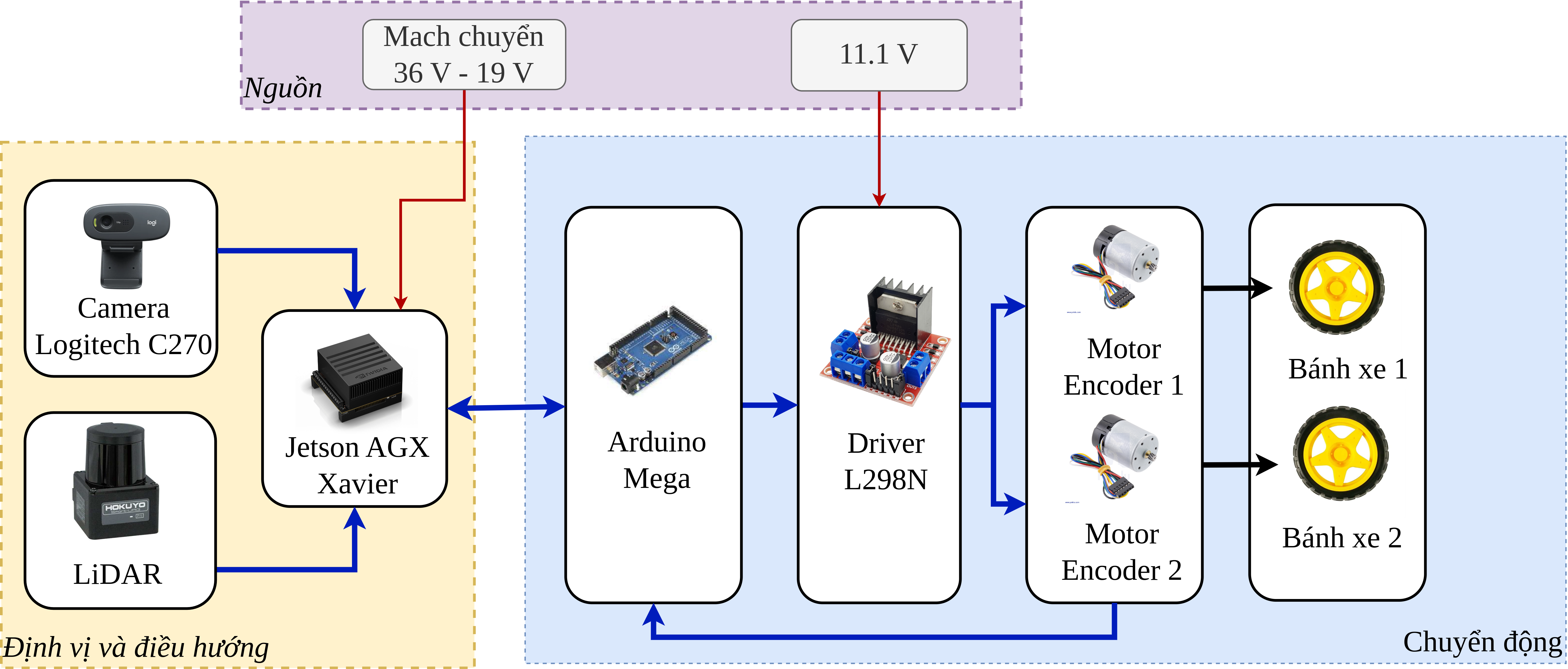}
    \caption{Sơ đồ kết nối các thành phần hệ thống}
    \label{fig:compoment}
\end{figure*}
Hệ điều hành ROS (Robot Operating System) là một framework phổ biến nhất trong công nghệ robot ngày nay \cite{thanh2021}. ROS cung cấp các bộ công cụ giả lập, trực quan hóa đồng thời trừu tượng hóa hệ thống phần cứng cũng như hỗ trợ điều khiển robot một cách dễ dàng. Hơn nữa, ROS có tập thư viện phong phú bao gồm các thuật toán SLAM, thuật toán điều hướng, .. từ đó giảm thiểu đáng kể chi phí, công sức trong việc xây dựng và phát triển robot.

\subsection{Các nghiên cứu liên quan}
Vấn đề định vị và điều hướng trong robot di động đã được các nhà nghiên cứu quan tâm từ trong và ngoài nước. Nghiên cứu \cite{David2010} sử dụng phương pháp Fuzzy nhằm định vị robot trong đó tập mờ được định nghĩa trong không gian vị trí của robot. Tuy nhiên phương pháp này không thể giải quyết được vấn đề kid-napping do vậy phương pháp định vị nổi tiếng là Monte Carlo (MCL) được xây dựng nhằm khắc phục vấn đề này. Các nghiên cứu  dựa trên bộ lọc Particle \cite{Vahid2018} hay bộ lọc Kalman và các biến thể của nó \cite{Khatib2015, Song2017} giải quyết vấn đề định vị cục bộ và toàn cục sử dụng kết hợp đa cảm biến nhằm giảm thiểu sai số định vị một cách đáng kể.

Các nghiên cứu \cite{Huijuan2015, Wang2016} sử dụng các điểm mốc đánh dấu nhằm cung cấp các tư thế tham chiếu toàn cục từ đó tư thế của robot được ước lượng thông qua mối quan hệ vị trí giữa điểm đánh dấu và robot.  Gần đây, hướng tiếp cận mới dựa trên mạng nơ-ron sâu được triển khai nhằm cải thiện khả năng bản địa hóa của robot \cite{Kendall2015}. Tuy nhiên các phương pháp này có độ chính xác thấp hơn so với phương pháp tiếp cận thông tin trước đó  như phương pháp hình học \cite{Sattler2019}.

Hướng tiếp cận dựa trên SLAM cũng được khai thác tiêu biểu như các nghiên cứu \cite{Han2018, Cristopher2018, Guan2021}. Nghiên cứu \cite{Xuexi2019} phân tích hệ thống định vị và điều hướng dựa trên SLAM từ đó cho thấy mức độ khả dĩ của hệ thống. Nghiên cứu \cite{Jesus2021} so sánh các thuật toán vSLAM cho kết quả ORB\_SLAM2 có độ chính xác về khoảng cách quỹ đạo cao trong nhà.
\subsection{Đóng góp của bài báo}
Trong bài báo này, chúng tôi đề xuất hệ thống định vị và dẫn đường cho robot hai bánh vi sai hoạt động trong môi trường trong nhà dựa trên monocular SLAM. Thuật toán ORB\_SLAM3 được khai thác nhằm sử dụng thông tin tư thế của camera thay thế cho các thông tin cơ bản của robot như Odometry, IMU nhằm cải thiện độ chính xác trong quá trình định vị. Chúng tôi thiết kế robot bao gồm máy tính nhúng hiệu năng cao Jetson Xavier AGX, camera 2D và LiDAR. Robot có khả năng di chuyển, thu thập dữ liệu để lập kế hoạch đường đi. Cuối cùng, hệ thống được trực quan hóa thông qua trình mô phỏng Gazebo, Rviz.
\subsection{Bố cục bài báo}
Bài báo được trình bày trong 5 phần: Phần \ref{Sec:intro} giới thiệu về bối cảnh nghiên cứu, đưa ra các nghiên cứu liên quan đồng thời nêu bật lên đóng góp của bài báo. Phần \ref{Sec:overview} mô tả tổng quan hệ thống, đưa ra mô hình phần cứng, mô hình động học của robot. Phương pháp đề xuất được trình bày trong phần \ref{Sec:System}. Phần \ref{Sec:result} đưa ra một số kết quả thử nghiệm nhằm đánh giá hiệu quả hoạt động của phương pháp đề xuất. Cuối cùng là kết luận và các hướng phát triển tiếp theo trong phần \ref{Sec:Conclusion}.
\section{TỔNG QUAN HỆ THỐNG}
\label{Sec:overview}
\subsection{Hệ thống phần cứng}
Hình \ref{fig:system_overview} mô tả tổng quan thành phần chính của hệ thống được tích hợp một mono camera (Logitech C270), một cảm biến LiDAR (Hokuyo 04-LX). Logitech camera thu nhận thông tin ảnh RGB, LiDAR Hokuyo thu thập thông tin từ môi trường sau đó chuyển tới máy tính nhúng Jetson Xavier AGX. Robot được điều khiển bằng máy tính khác thông qua wifi.

Sơ đồ kết nối các thành phần chi tiết của robot nhằm thực hiện cho việc di chuyển, định vị và điều hướng trong môi trường trong nhà được thể hiện qua Hình \ref{fig:compoment}. Các thành phần chia làm 3 thành phần chính: Phần định vị và điều hướng, phần chuyển động và phần nguồn.
 Máy tính nhúng Jetson Xavier AGX sử dụng ROS trên Ubuntu 18.04 đóng vai trò làm bộ xử lý trung tâm, tiếp nhận thông tin từ các node bao gồm Camera, LiDAR, Arduino Mega sau đó phân bổ ngược trở lại. Bộ phận chuyển động bao gồm vi điều khiển Arduino Mega, driver điều khiển L298 và 2 động cơ DC. Khối nguồn cung cấp điện áp cho hai khối trên bao gồm một bộ nguồn 11.1 V cho khối chuyển động và bộ chuyển đổi từ 36V xuống 19V cho khối định vị và điều hướng. Robot được thiết kế nhỏ gọn đảm bảo tính an toàn và hiệu quả của thiết bị.

\subsection{Mô hình động học của robot}
\begin{figure}[!ht]
    \centering
    \includegraphics[width=0.4\textwidth]{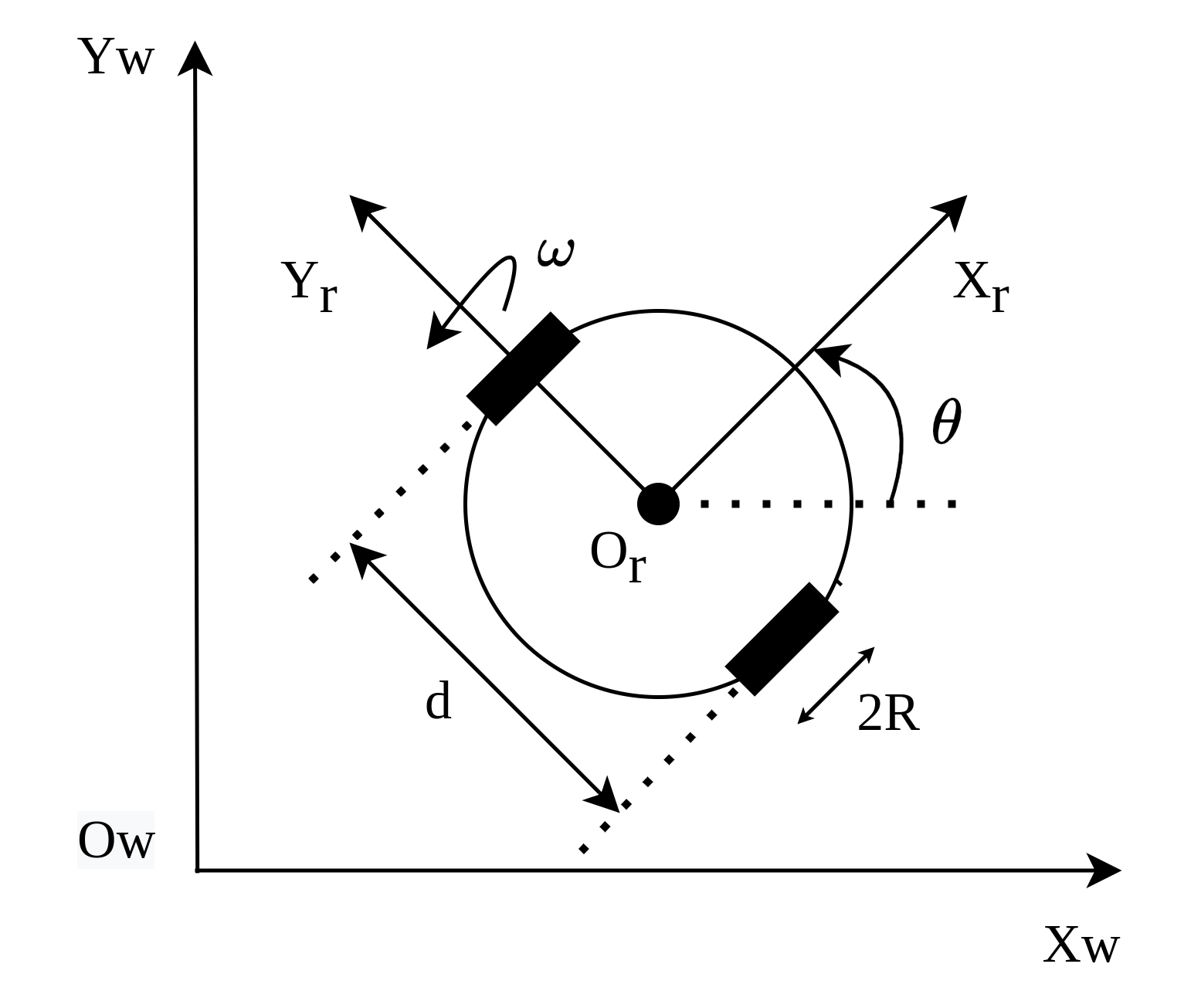}
    \caption{Hệ trục tọa độ của robot hai bánh vi sai}
    \label{fig:kinematic}
\end{figure}

Chúng tôi sử dụng robot hai bánh vi sai trong mặt phẳng đất như Hình \ref{fig:kinematic}. Tư thế của robot được xác định trong mặt phẳng $(X_w, Y_w)$ với vị trí $(x, y)$ và hướng $\theta$. Phương trình động học của robot được cho bởi:
\begin{equation}
\label{Eq:kinematic_Robot}
\begin{aligned}
& \dot{x} = vcos (\theta)\\
& \dot{y} = vsin (\theta)\\
& \dot{\theta} = \omega
\end{aligned}
\end{equation}
trong đó $v$, $\omega$ lần lượt là vận tốc tuyến tính và vận tốc góc của robot. Do robot được điều khiển bởi máy tính nhúng, chúng ta cần rời rạc hóa phương trình động học thuận. Gọi $\Delta{t}$ là thời gian lấy mẫu và $(x_t, y_t, \theta_t) $là tư thế của robot tại bước thời gian $t$. Đạo hàm có thể được tính gần đúng như sau:

\begin{equation}
\begin{aligned}
\dot{x} = \frac{x_{t_1 - x}}{\Delta{t}}
\end{aligned}
\end{equation}

Phương trình động học thuận rời rạc được đưa ra bởi:
\begin{equation}
\begin{aligned}
& x_{t+1} = x_t + v_t \Delta t cos (\theta_t)\\
& y_{t+1} = y_t + v_t \Delta t sin (\theta_t)\\
& \theta_{t+1} = \theta_t + \omega_t \Delta t
\end{aligned}
\label{Eq:kinematicRobot}
\end{equation}

\section{ĐỊNH VỊ VÀ ĐIỀU HƯỚNG DỰA TRÊN MONOCULAR SLAM}
\label{Sec:System}
\begin{figure*}[!ht]
    \centering
    \includegraphics[width=0.8\textwidth]{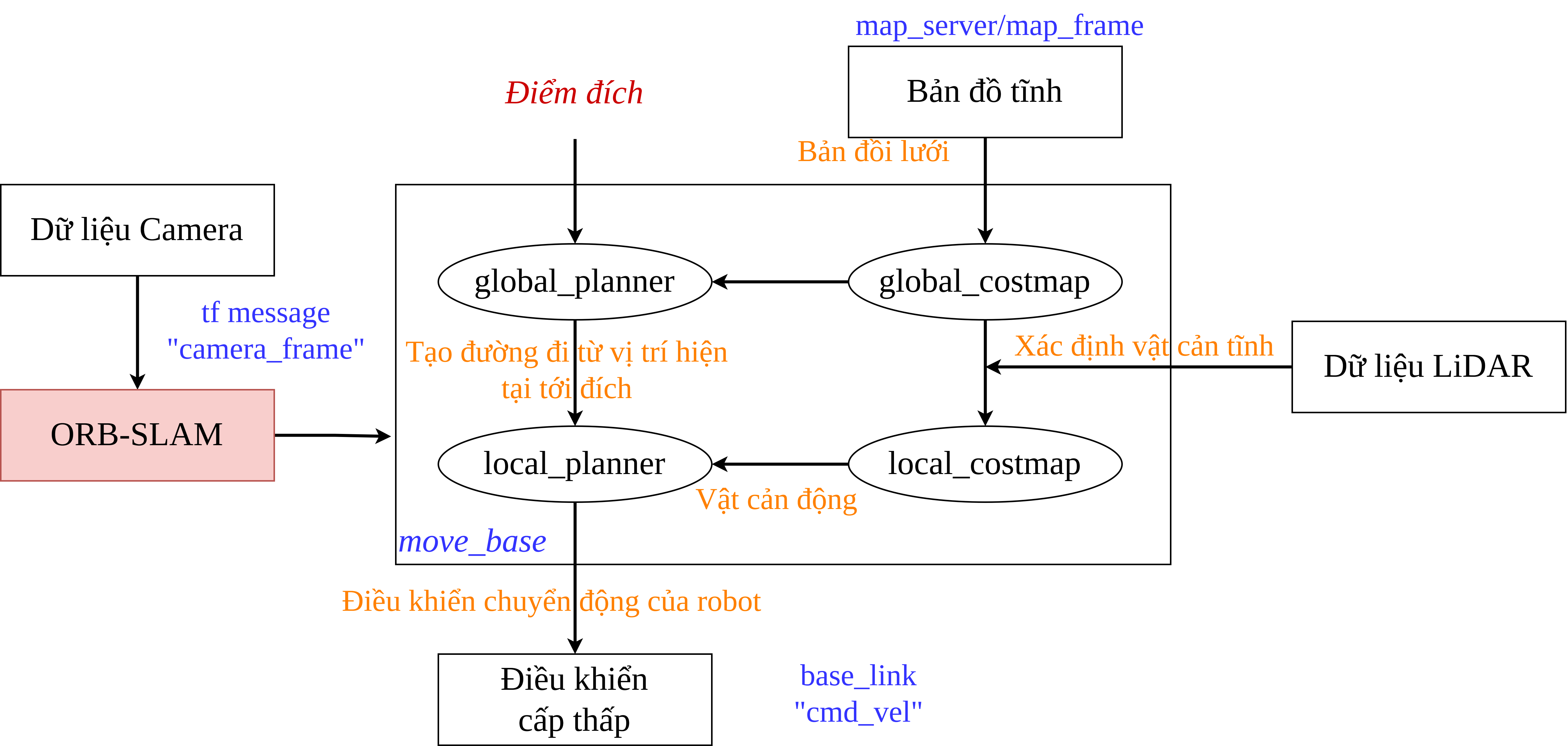}
    \caption{Tổng quan hệ thống định vị và dẫn đường dựa trên monocular SLAM}
    \label{fig:propose}
\end{figure*}

Chúng tôi để xuất triển khai ORB\_SLAM vào trong Navigation Stack của ROS đề điều hướng cho robot. Khối AMCL là một khối tùy chọn được Navigation Stack cung cấp, chúng tôi thay thế bằng phương pháp đề xuất của chúng tôi dựa trên ORB\_SLAM nhằm định vị robot dựa trên monocular camera. Các thành phần trong kiến trúc ROS chúng tôi đề xuất dựa trên navigation stack và monocular slam được thể hiện như trong Hình \ref{fig:propose}.

ORB\_SLAM \cite{Campos_2021} sử dụng ảnh RGB làm đầu vào. Tại mỗi khung hình, một tập các điểm đặc trưng được trích xuất . Các khung hình chính được lựa chọn dựa trên khả năng đồng hiển thị với khung hình khác. Các khung hình chính và các điểm đặc trưng được lưu trữ để thực hiện bản địa hóa và chọn các khung hình mới. Các phép dịch chuyển của camera được tính toán (vị trí, góc) trong hệ thống sau đó được chuyển qua hệ tọa độ của robot từ đó xác định được tư thế của robot trong bản đồ. Hệ thống định vị và điều hướng chúng tôi đề xuất sử dụng ORB\_SLAM nhằm ước tính tư thế của robot và LiDAR cho tác vụ phát hiện vật cản xung quanh, dữ liệu bản đồ tĩnh cho lập kế hoạch chuyển động. 

Gói ROS \textit{move\_base} được sử dụng  nhằm thực hiện điều hướng tự động. Gói này sử dụng thông tin bản địa hóa từ cảm biến và cung cấp lệnh di chuyển tới robot để robot di chuyển an toàn trong môi trường mà không va chạm với chướng ngại vật. Hệ thống điều hướng cho robot bao gồm hai phần chính: Phần lập quỹ đạo toàn cục (\textit{global\_planner}) và lập quỹ đạo cục bộ (\textit{local\_planner}). Công cụ lập quỹ đạo toàn cục dựa dựa trên thuật toán tìm kiếm A$^*$ cho phép robot sử dụng thông tin bản đồ tĩnh từ môi trường từ đó lập quỹ đạo toàn cục đảm bảo di chuyển tới đích với quỹ đạo nhắn nhất và an toàn tránh các vật thể tĩnh mà không cần xem xét tới các ràng buộc về động học của robot. Công cụ lập kế hoạch cục bộ dựa trên thuật toán DWA sử dụng quỹ đạo tham chiếu thu được từ quỹ đạo toàn cục từ đó tạo ra các lệnh vận tốc dựa trên tư thế của robot được tính toán qua ORB\_SLAM và đảm bảo rằng robot có thể di chuyển tới đích. Bên cạnh đó, hai bản đồ chi phí (\textit{cost\_map}) được cung cấp cho cả lập bản đồ toàn cục và lập bản đồ cục bộ nhằm mở rộng phạm vi của vật cản, tránh va chạm trong quá trình di chuyển.

Trong quá trình điều hướng tự động, các chướng ngại vật động có thể được phát hiện bằng 2D LiDAR. Thông qua việc cập nhật bản đồ chi phí cục bộ (\textit{local\_costmap}) để nhận ra khả năng tránh chướng ngại vật động. Ước tính tư thế ORB\_SLAM chính xác góp phần khởi tạo cho việc lập  quỹ đạo toàn cục trong khi ước tính tư thế mạnh mẽ và theo thời gian thực của ORB\_SLAM giúp lập quỹ đạo cục bộ một cách chính xác.

\section{KẾT QUẢ}
\label{Sec:result}
Hiệu suất của hệ thống định vị và vẽ bản đồ được đánh giá thông qua phầm mềm mô phỏng Gazebo và trực quan hóa qua phầm mềm Rviz. 

\subsection{Thiết lập môi trường}
\begin{figure}[!ht]
    \centering
    \includegraphics[width=0.4\textwidth]{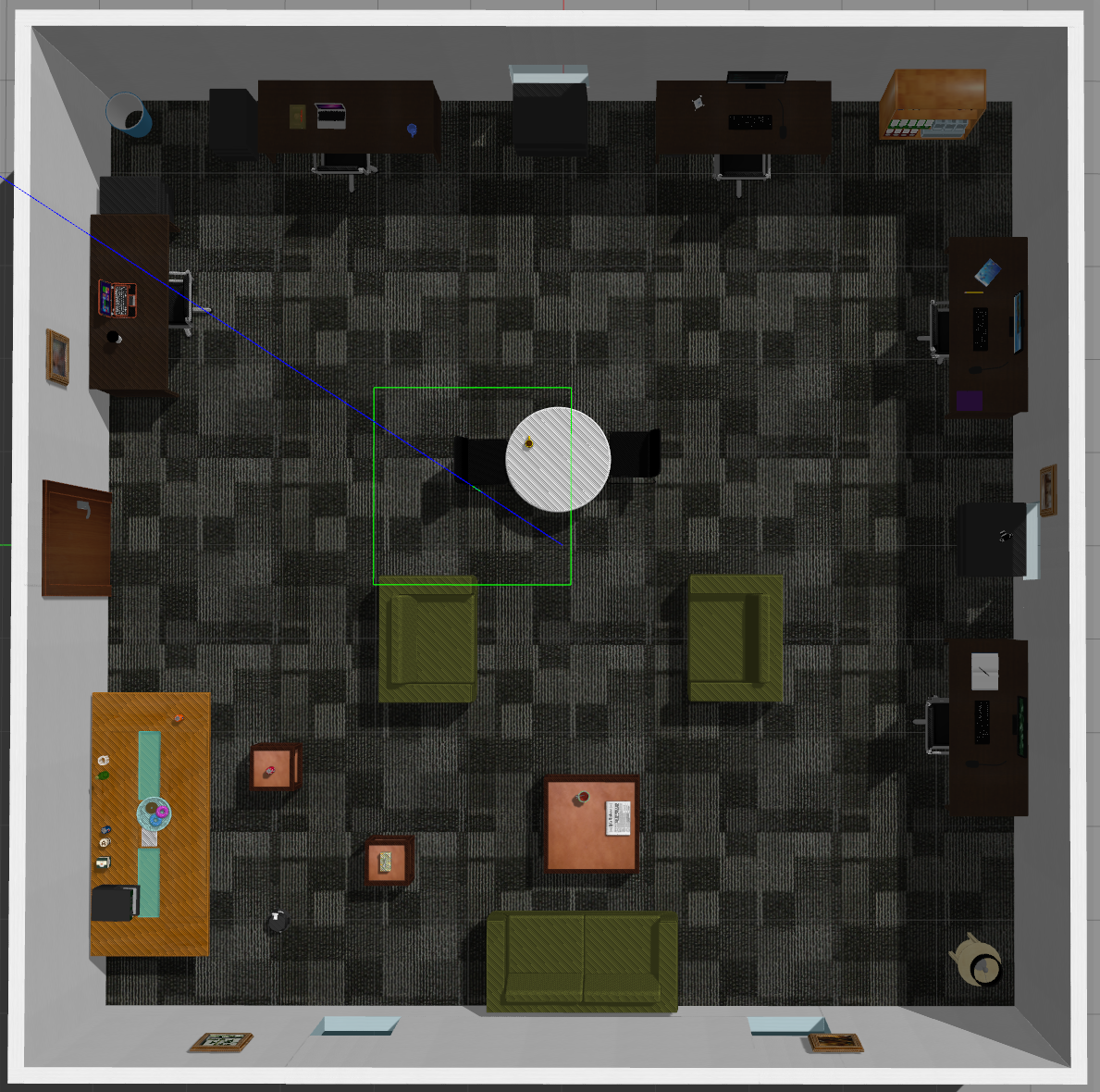}
    \caption{Môi trường giả lập trong nhà}
    \label{fig:simulation}
\end{figure}

Chúng tôi xây dựng môi trường giả lập trong nhà bao gồm nhiều đồ dùng, bàn ghế, .. sao cho giống môi trường thực tế nhất được thể hiện như trong Hình \ref{fig:simulation}. Môi trường được bao quanh bởi các vách tường và các vật thể được sắp xếp ngẫu nhiên trên môi trường.

\begin{figure}[!ht]
    \centering    \includegraphics[width=0.3\textwidth]{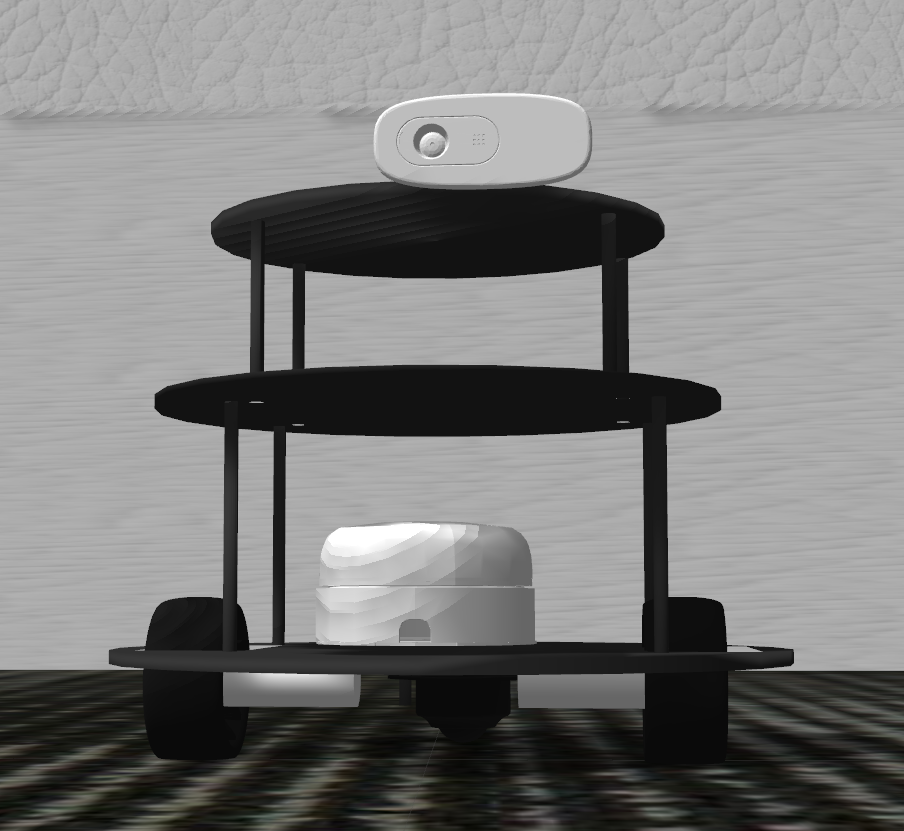}
    \caption{Mô hình 3D robot hai bánh vi sai}
    \label{fig:robot}
\end{figure}

Mô hình robot được thiết kế như trong Hình \ref{fig:robot} với các tham số mô phỏng và thực nghiệm giống nhau được thể hiện qua Bảng \ref{tab:specRobot}. Robot được tối giản hóa trong môi trường mô phỏng, chỉ giữ lại những thành phần cơ bản cho định vị và dẫn đường bao gồm khung xe, động cơ, bánh xe, cảm biến LiDAR và camera 2D. Thông số cơ bản cho cảm biến LiDAR và camera 2D lần lượt được mô tả chi tiết trong Bảng \ref{tab:specLidar} và Bảng \ref{tab:specCamera}.

\begin{table}[]
    \centering
    \caption{Thông số mô hình robot}
    \begin{tabularx}{0.453\textwidth}{
    | >{\centering\arraybackslash}m{0.2\textwidth} 
    | >{\arraybackslash}m{0.2\textwidth}|
    }
    \hline
    \bfseries Tốc độ tuyến tính & 0.30 (m/s) \\\hline
    \bfseries Tốc độ góc & 0.20 (rad/s)\\\hline
    \bfseries Bán kính robot & 0.22 (m)\\\hline
    \bfseries Bán kính bánh xe & 0.03 (m)\\\hline
    \end{tabularx}
    \label{tab:specRobot}
\end{table}

\begin{table}[]
    \centering
    \caption{Thông số cảm biến LiDAR}
    \begin{tabularx}{0.453\textwidth}{
    | >{\centering\arraybackslash}m{0.2\textwidth} 
    | >{\arraybackslash}m{0.2\textwidth}|
    }
    \hline
    \bfseries Góc quét & 270 ($^\circ$) \\\hline
    \bfseries Tần số quét & 10 (Hz)\\\hline
    \bfseries Phạm vi & 0.06 - 6.40 (m)\\\hline
    \bfseries Độ phân giải góc & 0.36 ($^\circ$) ($360 ^\circ / 1024$)\\\hline
    \end{tabularx}
    \label{tab:specLidar}
\end{table}

\begin{table}[]
    \centering
    \caption{Thông số camera 2D}
    \begin{tabularx}{0.453\textwidth}{
    | >{\centering\arraybackslash}m{0.2\textwidth} 
    | >{\arraybackslash}m{0.2\textwidth}|
    }
    \hline
    \bfseries Độ phân giải RGB & 1280 x 720 (pixel) \\\hline
    \bfseries Giới hạn khoảng cách & 0.02 - 20 (m)\\\hline
    \bfseries Tốc độ khung hình & 30 (FPS)\\\hline
    \bfseries Góc quét & 270 ($^\circ$) \\\hline
    \end{tabularx}  
    \label{tab:specCamera}
\end{table}

\subsection{Kết quả mô phỏng}
\begin{figure}[!ht]
    \centering
    \includegraphics[width=0.3\textwidth]{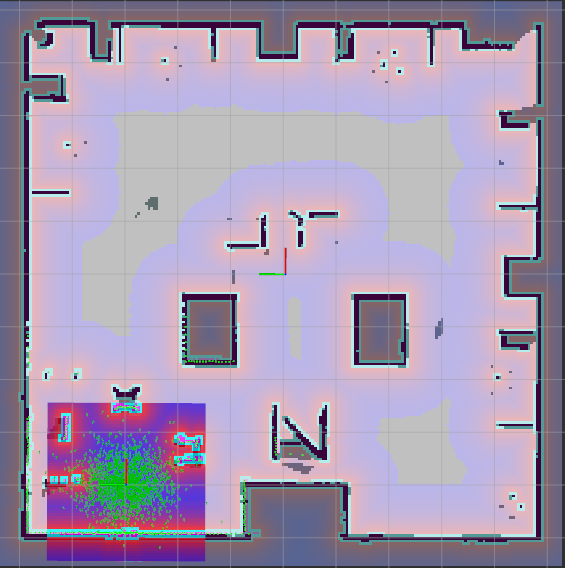}
    \caption{Trực quan hóa bản đồ  tĩnh của môi trường}
    \label{fig:rviz}
\end{figure}

Đầu tiên, robot được khởi tạo trong môi trường như trong Hình \ref{fig:simulation}. Tiếp theo chúng tôi thực hiện quét bản đồ môi trường nhằm thu thập bản đồ tĩnh của môi trường thông qua Google CartoGrapher. Sau đó, bản đồ tĩnh được thêm vào khối cost\_map và khối move\_base nhằm triển khai điều hướng được trực quan hóa trong Rviz như trong Hình \ref{fig:rviz}. Bản đồ tĩnh sau khi đi qua khối cost\_map bao gồm các ô có giá trị từ 0 đến 255 trong đó các khối màu đen biểu diễn các vùng chứa vật cản, các cùng màu trắng biểu thị khu vực trống, vùng màu xám biểu thị khu vực chưa biết rõ. Quanh mỗi khu vực vật cản được mở rộng thêm một vùng (màu đỏ) nhằm tránh việc robot đụng chạm trong quá trình di chuyển.

\begin{figure}[!ht]
    \centering
    \captionsetup{justification=centering}
    \includegraphics[width=0.4\textwidth]{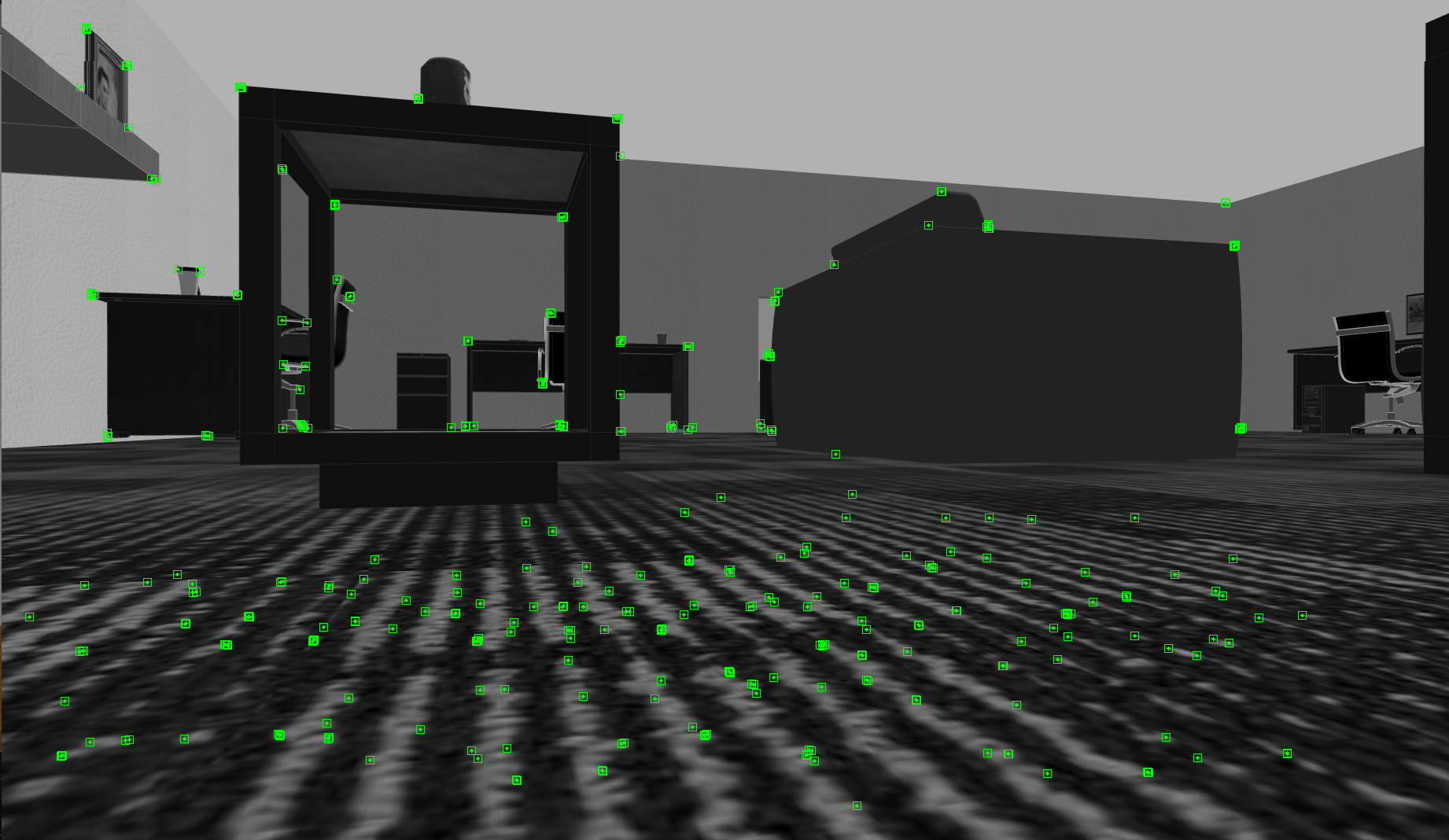}
    \caption{Trích xuất đặc trưng của ảnh thông qua ORB\_SLAM}
    \label{fig:keyframe}
\end{figure}

\begin{figure}[!ht]
    \centering
    \includegraphics[width=0.4\textwidth]{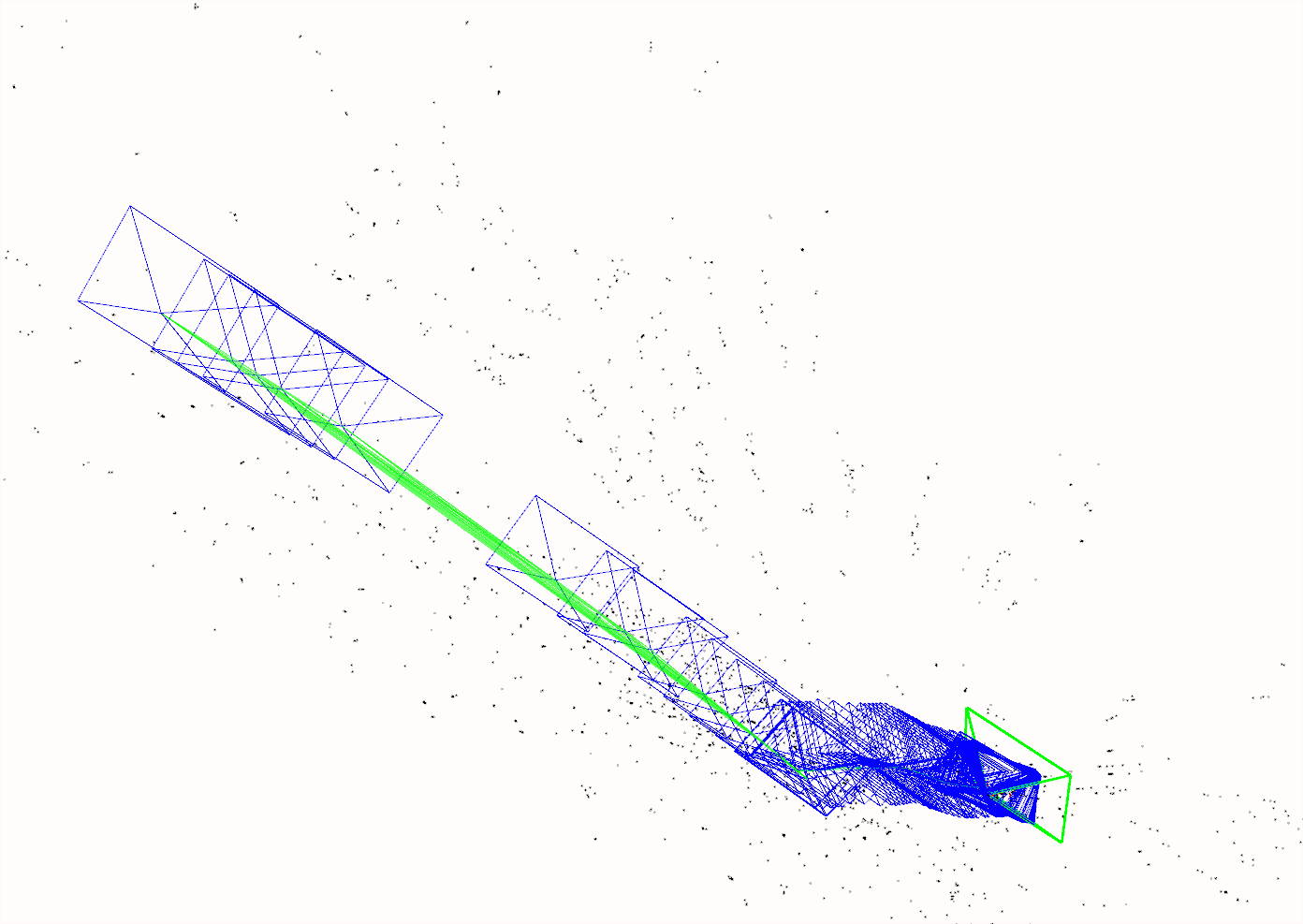}
    \caption{Tư thế của camera thông qua ORB\_SLAM}
    \label{fig:orbpose}
\end{figure}

Hệ thống định vị được thiết lập dựa trên ORB\_SLAM trong đó nhận thông tin đầu vào từ ảnh RGB của camera. Hình \ref{fig:keyframe} thể hiện các đặc trưng của ảnh được trích xuất thông qua thuật toán được biểu diễn bởi các điểm màu xanh. Tư thế của robot dựa trên mối tương quan giữa các điểm đặc trưng của khung hình trước và khung hình sau từ đó tạo thành quỹ đạo chuyển động như trong Hình \ref{fig:orbpose}.

\begin{figure}[!ht]
    \centering
    \captionsetup{justification=centering}
    \includegraphics[width=0.3\textwidth]{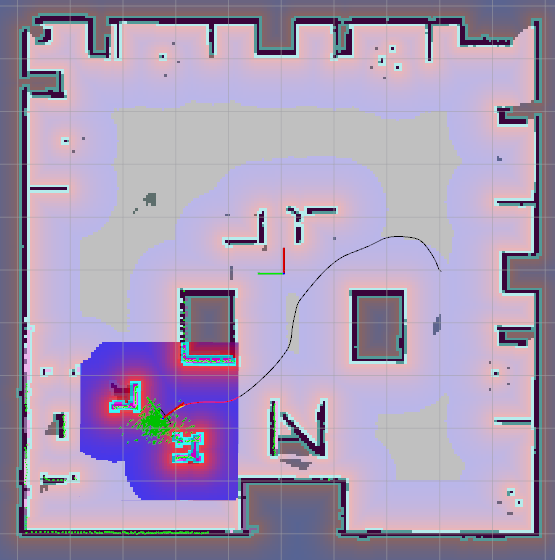}
    \caption{Quỹ đạo chuyển động của robot trong quá trình điều hướng}
    \label{fig:navigation}
\end{figure}

\begin{figure*}[!ht]
     \centering
     \begin{subfigure}[b]{0.3\textwidth}
         \centering
         \includegraphics[width=\textwidth]{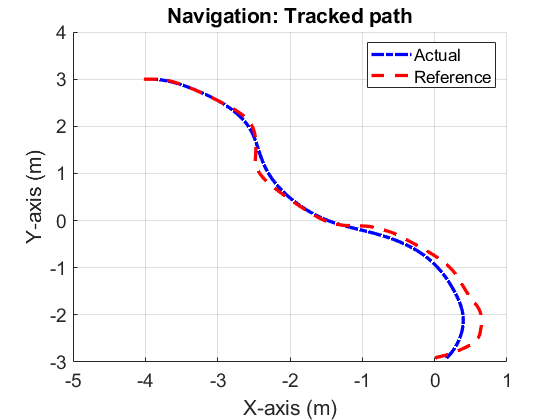}
         \caption{Quỹ đạo bám của các bộ điều khiển}
         \label{fig:trajectory}
     \end{subfigure}
     \hfill
     \begin{subfigure}[b]{0.3\textwidth}
         \centering
         \includegraphics[width=\textwidth]{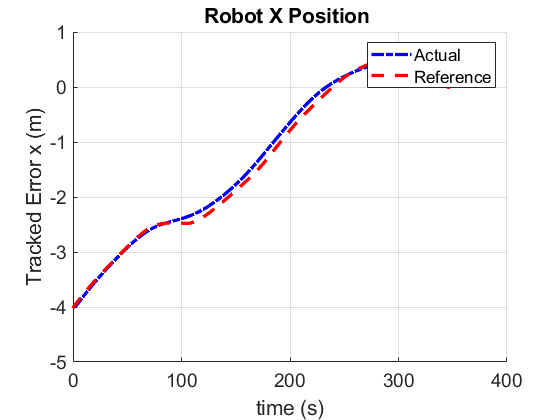}
         \caption{Sai số theo trục x}
         \label{fig:errorX}
     \end{subfigure}
     \hfill
     \begin{subfigure}[b]{0.3\textwidth}
         \centering
         \includegraphics[width=\textwidth]{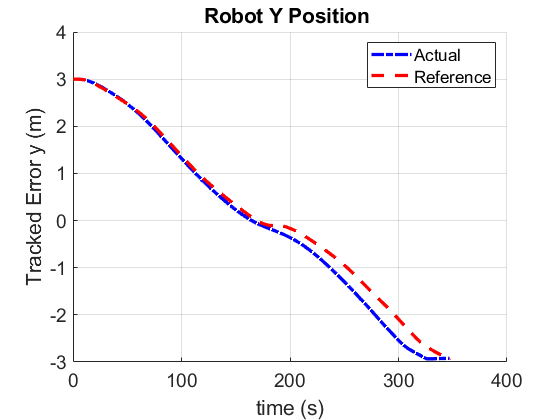}
         \caption{Sai số theo trục y}
         \label{fig:errorY}
     \end{subfigure}
        \caption{Sai số điều hướng theo từng thành phần}
        \label{fig:errorTrj}
\end{figure*}

Tiếp theo, Robot được di chuyển từ A (-4.0, 3.0) tới B (0.2, -3.0) theo tọa độ (x,y). Chúng tôi sử dụng thuật toán A$^*$ nhằm tính toán quỹ đạo toàn cục cho robot và thuật toán DWA nhằm tính toán quỹ đạo cục bộ. Kết quả của quá trình điều hướng được thể hiện qua Hình \ref{fig:navigation}. Robot được định vị thông qua ORB\_SLAM, các thông tin về môi trường được thu thập thông qua cảm biến LiDAR. Hệ thống điều hướng sẽ bao gồm hai loại quỹ đạo. Quỹ đạo toàn cục nhằm đưa ra quỹ đạo sơ bộ cho robot là đường ngắn nhất và an toàn để robot có thể di chuyển từ điểm xuất phát tới điểm đích. Quỹ đạo cục bộ sẽ tính toán lại và đưa ra quỹ đạo đã được điều chỉnh phù hợp cho robot để tránh được vật cản theo từng thời điểm thông qua đường quỹ đạo màu đỏ.

Sai số của quỹ đạo tham chiếu và vị trí thực tế của robot được thể hiện thông qua Hình \ref{fig:errorTrj}. Hệ thống có độ chính xác cao với sai số nhỏ hơn 0.25m trong đó sai số trong từng thành phần x, y lần lượt là nhỏ hơn 0.1m và 0.3m. Một phần sai số kết quả dựa trên việc thực hiện trích xuất đặc trưng khó khăn trong môi trường giả lập hơn thực tế.
\section{KẾT LUẬN}
\label{Sec:Conclusion}
Bài báo đề xuất hệ thống định vị và điều hướng cho robot hai bánh vi sai trong nhà dựa trên hệ điều hành cho robot ROS dựa trên monocular SLAM. Phương pháp monocular SLAM cho phép robot thu thập, tiếp nhận và xử lý thông tin từ môi trường nhằm định vị vị trí hiện tại trên bản đồ. Sau khi định vị được vị trí, hệ thống điều hướng tính toán quỹ đạo cục bộ và toàn cục dựa trên thuật toán DWA và A$^*$. Các kết quả mô phỏng, thử nghiệm cho thấy tính hiệu quả của hệ thống định vị và điều hướng cho robot di động trong nhà.
%

\bibliographystyle{IEEEtran}
\balance
\bibliography{reference}
\end{document}